\pdfoutput=1
\documentclass{article}
\usepackage{spconf,amsmath,graphicx}
\usepackage{amsfonts}

\usepackage{subfig}
\usepackage[colorlinks,linkcolor=blue]{hyperref}

\def\L{{\cal L}}

\title{ATTRIBUTE-CONTROLLED FACE PHOTO SYNTHESIS FROM SIMPLE LINE DRAWING}
\name{Qi Guo \ \ Ce Zhu \ \ Zhiqiang Xia \ \ Zhengtao Wang \ \ Yipeng Liu\thanks{This research is supported by National Natural Science Foundation of China (NSFC, No. 61571102, No. 61602091), the Fundamental Research Funds for the Central Universities (No. ZYGX2014Z003) and National High Technology Research and Development Program of China (863, No.2015AA015903). Portions of the research in this paper use the ZJU-VIPA Line Drawing Face Database collected under the joint sponsor of National Natural Science Foundation of China and Zhejiang Natural Science Foundation Program.}}
\address{School of Electronic Engineering / Center for Robotics\\
University of Electronic Science and Technology of China (UESTC), Chengdu, China
}
\begin{document}
%
\maketitle
\begin{abstract}
Face photo synthesis from simple line drawing is a one-to-many task as simple line drawing merely contains the contour of human face. Previous exemplar-based methods are over-dependent on the datasets and are hard to generalize to complicated natural scenes. Recently, several works utilize deep neural networks to increase the generalization, but they are still limited in the controllability of the users. In this paper, we propose a deep generative model to synthesize face photo from simple line drawing controlled by face attributes such as hair color and complexion. In order to maximize the controllability of face attributes, an attribute-disentangled variational auto-encoder (AD-VAE) is firstly introduced to learn latent representations disentangled with respect to specified attributes. Then we conduct photo synthesis from simple line drawing based on AD-VAE. Experiments show that our model can well disentangle the variations of attributes from other variations of face photos and synthesize detailed photorealistic face images with desired attributes. Regarding background and illumination as the style and human face as the content, we can also synthesize face photos with the target style of a style photo.
\end{abstract}
\section{Introduction}
\label{sec:intro}

\begin{figure*}[htbp]
\centering
   \includegraphics[width=420pt,keepaspectratio]{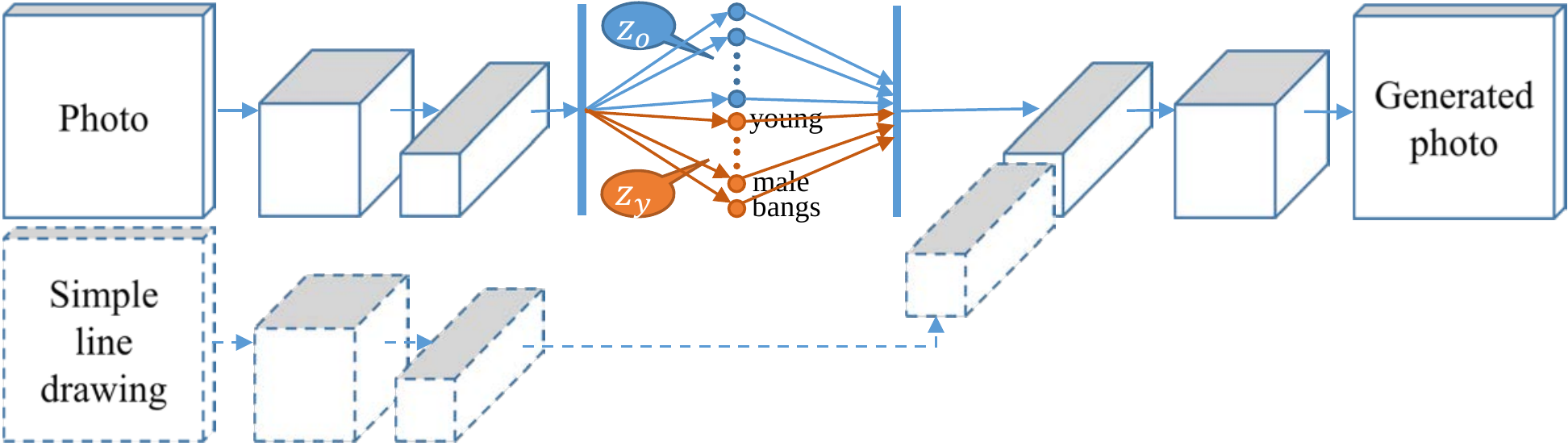}
   \caption{Model architecture. Solid lines depict the basic structure of AD-VAE. Dashed lines show another channel of convolutional nueral network which takes simple line drawing as input.}
   \label{fig:model}
\end{figure*}

Face sketch-photo synthesis has been well developed in recent years for its widely application on law enforcement. An experienced artist draws a sketch about the suspect according to the description of the witness. Then the sketch is transformed to a photorealistic face image. At last, the synthesized face photo is compared to the images from face datasets to find the suspect.

The majority of prior works utilize sketches with rich details that are difficult to draw. Liang et al. \cite{liang2012face} conduct face sketch-photo synthesis from simple line drawing. As a rough kind of sketch, simple line drawing, merely containing the contour of human face, is easy to be obtained and modified for ordinary people, which also makes photo synthesis from simple line drawing a quite tough task. In most cases, witnesses can remember some visual attributes (hair colour, complexion, etc.) of suspects apart from the outline. We take advantage of these face attributes to add rich details to the outline of the face. A simple line drawing with different face attributes correspond to different faces.

The prior works of face sketch-photo synthesis can be divided into two categories: traditional methods and deep learning methods.

Most of traditional methods are based on an image patch dictionary of exemplars \cite{liang2012face,liu2007bayesian,xiao2009new,wang2009face,gao2012face,zhou2012markov,wang2013transductive,peng2016multiple}. With the number of training data grows and the complicacy of chosen representation of image patch increases, the testing time grows linearly. Moreover, these exemplar-based methods require both training and testing faces constrained under the same conditions, such as the same race, front pose and similar background. These restrictions hinder them from generalizing to more complicated natural scenes. Even though Peng et al. \cite{peng2016multiple} utilize multiple representations of image to avoid suffering from these limitations at the cost of test time consumption, they can scarcely synthesize natural face image with complex backgrounds and diverse poses.

Recently, deep learning is explosively used in computer vision for its powerful generalization in highly complex tasks. A trained deep neural network is quite fast in testing stage due to its feed-forward framework, which allows users to obtain the results in real time. Convolutional Sketch Inversion (CSI) \cite{guccluturk2016convolutional} and Scribbler \cite{sangkloy2016scribbler} apply deep learning into sketch-photo synthesis. The architecture in \cite{guccluturk2016convolutional} is a one-to-one convolutional nueral network, namely, it generates only one photo from one sketch. Scribbler \cite{sangkloy2016scribbler} makes use of generative adversarial networks (GANs) \cite{goodfellow2014generative,denton2015deep,radford2015unsupervised} conditioned on sketched boundaries and sparse color strokes to generate realistic faces, cars and bedrooms. It allows users to scribble over the sketch to indicate preferred color for objects. Complex as human faces, simple color strokes are not enough to generate rich details.


In this paper, we propose a deep generative model to synthesize face photo from simple line drawing controlled by face attributes. First, an attribute-disentangled variational auto-encoder (AD-VAE) is introduced to learn latent representations disentangled with respect to specified face attributes. We regard face attributes as some high-level variations of face images and specify a factored set of latent variables in variational auto-encoder (VAE) \cite{rezende2014stochastic,kingma2014variational} to capture these variations utilizing the binary attribute labels. The remaining latent variables are to learn other factors of variation such as illumination and background. Then we conduct face photo synthesis from simple line drawing by adding another channel of convolutional neural network to the AD-VAE which takes simple line drawing as input.

Different from attribute-conditioned variational auto-encoder (AC-VAE) \cite{yan2016attribute2image}, AD-VAE adds an inference from input images to the attribute variables, which can turn discrete variables to be continuous. This supervised manner is similiar to \cite{cheung2015discovering}, but we adopt a generative way. \cite{kulkarni2015deep} force some latent variables to specifically represent active transformations in 3D datasets by organizing their training data changing in only one single scene variable for each mini-batch. However, such data organization is unmanageable for many natural image databases.

Regarding background and illumination as the style and human face as the content, simple line drawing and face attibutes control the content of the generated face image, and the other latent variables control the style of the generated face image. Given a simple line drawing, the proposed method can generate different face photos with desired attributes and random styles. Given a simple line drawing and a style photo, we can also synthesize photo with target style.


\section{Methods}
\label{sec:Methods}

\subsection{Variational auto-encoder}
\label{subsec:VAE}

\begin{figure*}[t]
\centering

\subfloat[Mouth\_slightly\_open]{
	\label{fig:improved_subfig_b}
	\begin{minipage}[t]{163pt}
		\centering
		\includegraphics[height=5.7cm, keepaspectratio=True]{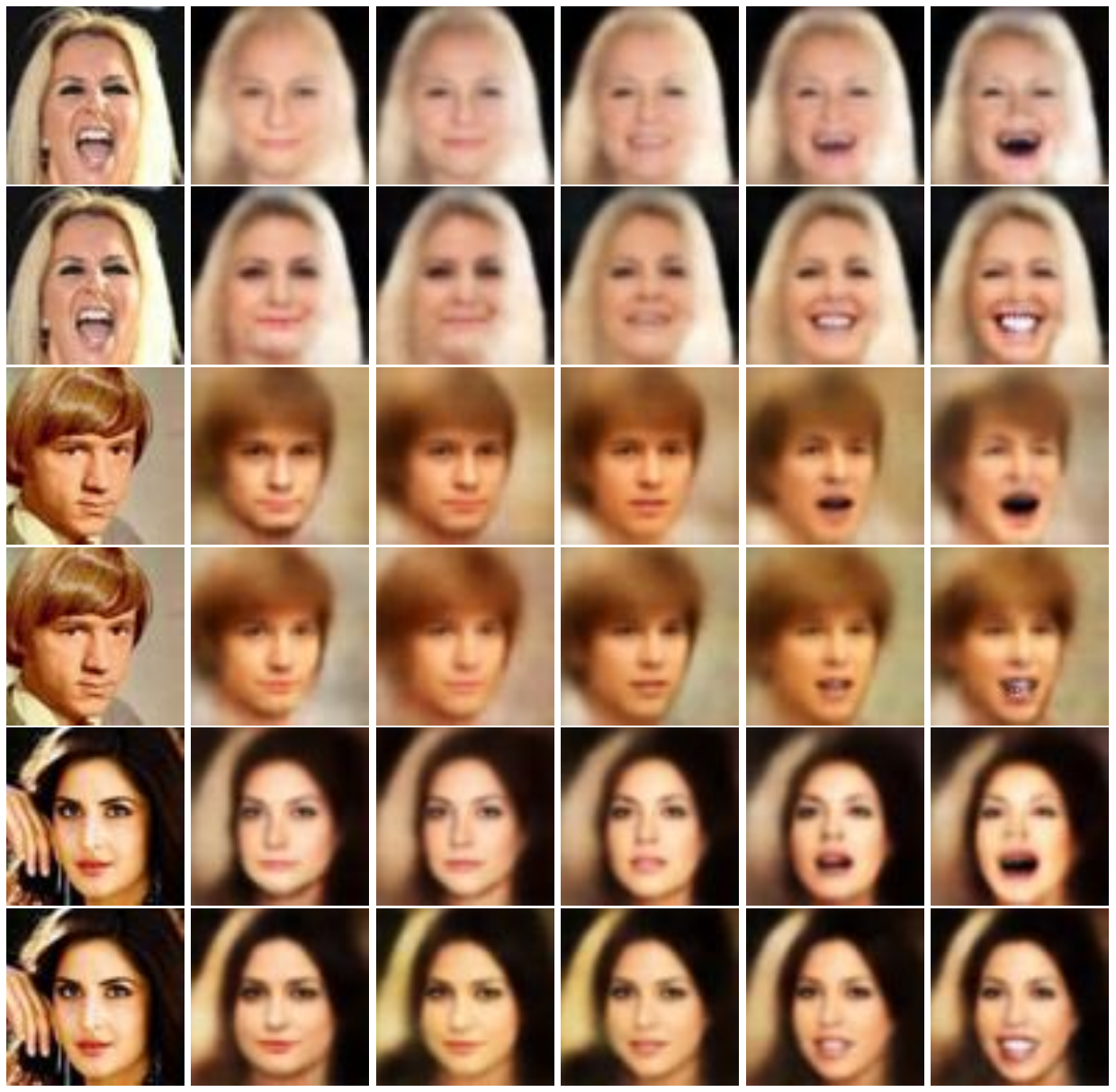}
	\end{minipage}
}
\subfloat[Eyeglasses]{
	\label{fig:improved_subfig_b}
	\begin{minipage}[t]{163pt}
		\centering
		\includegraphics[height=5.7cm, keepaspectratio=True]{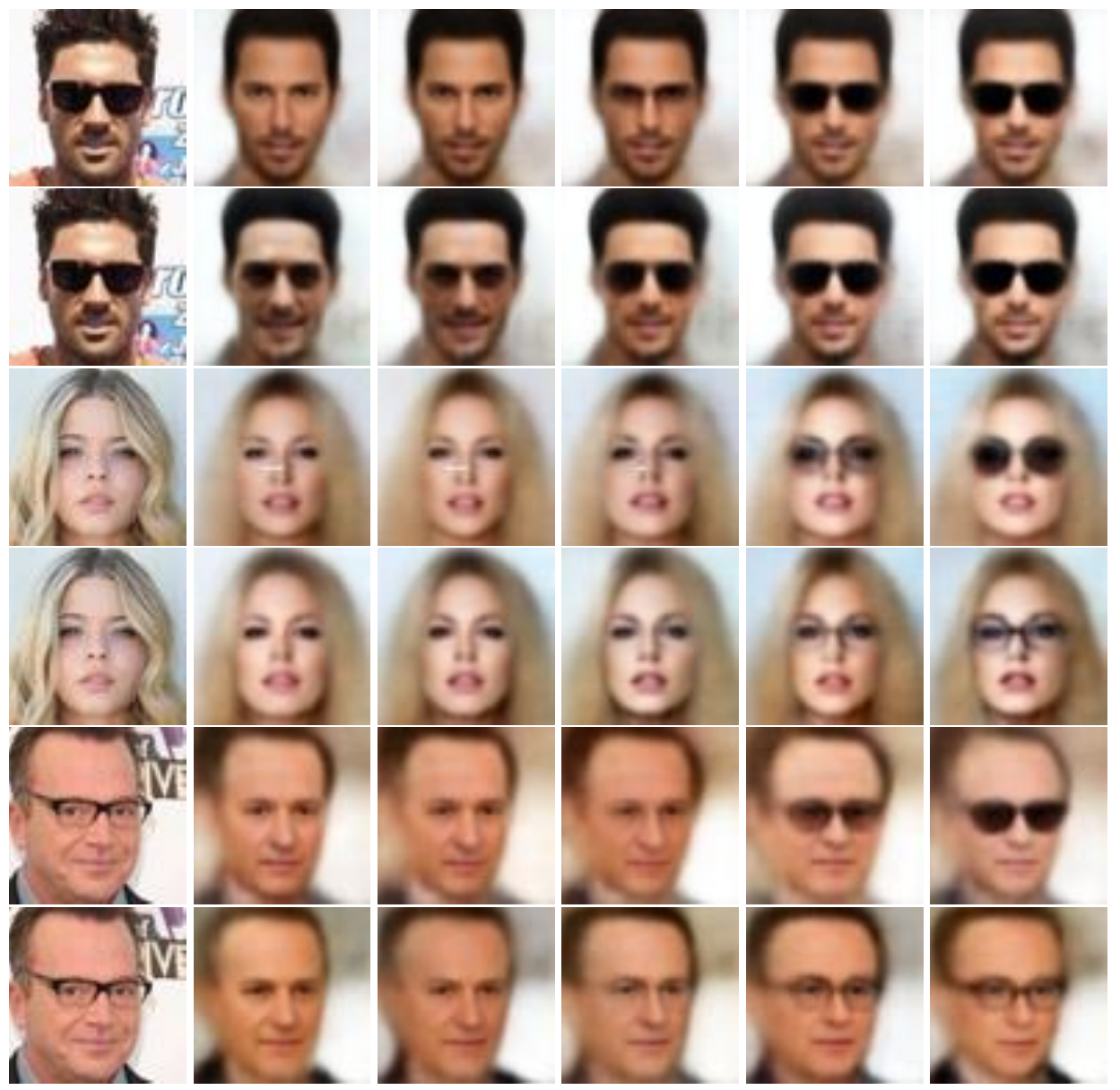}
	\end{minipage}
}
\subfloat[Bangs]{
	\label{fig:improved_subfig_a}
	\begin{minipage}[t]{163pt}
		\centering
		\includegraphics[height=5.7cm, keepaspectratio=True]{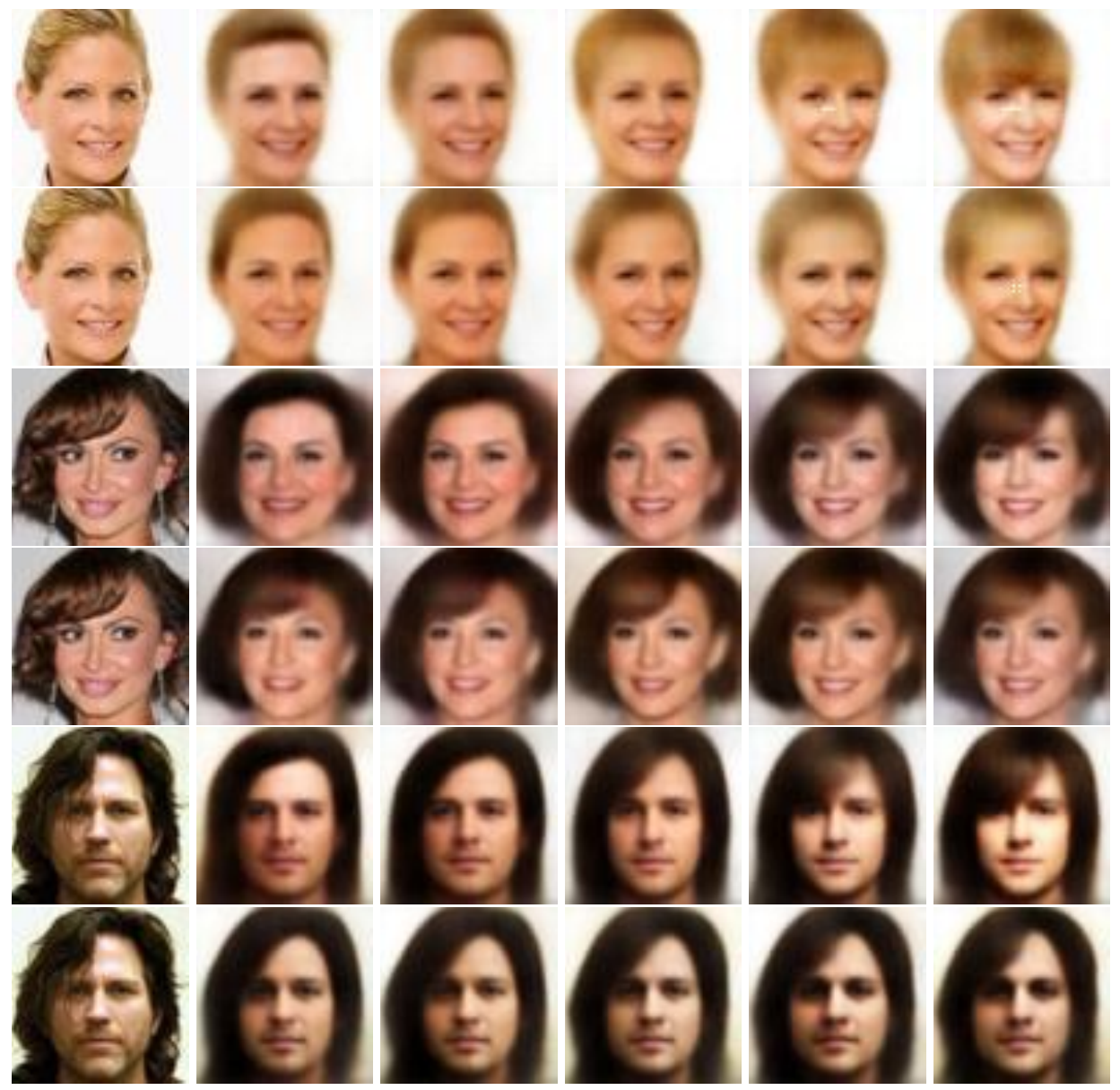}
	\end{minipage}
}

\caption{Manipulating attribute variables. For each group, the first column is ground truth. For each ground truth, the first row is generated by AD-VAE and the second row is gnerated by AC-VAE.}
\label{fig:attribute}
\end{figure*}

Given an input ${x} \in \mathbb{R}^N$ and its corresponding latent variables ${z} \in \mathbb{R}^M$, the basic structure of VAE consists of two networks: an encoder $q_{\phi}(z|x)$ (recognition model) to approximate posterior inference and a decoder $p_{\theta}(x|z)$ (generative model) to map the latent variables to data space. Due to the intractable posterior $p_{\theta}(x|z)$, we maximize the following variational lower bound of log-likelihood $logp_{\theta}(x)$:
\begin{equation}
	\label{eq:vae}
	\L(x;\phi,\theta)=-D_{KL}(q_{\phi}(z|x)||p_{\theta}(z))+E_{q_{\phi}(z|x)}[logp_\theta(x|z)]
\end{equation}
where $p_{\theta}(z)$ is the prior distribution of the latent variables $z$, generally a simple isotropic unit Gaussian. $D_{KL}$ is Kullback-Leibler divergence.

\subsection{Attribute-disentangled variational auto-encoder}
\label{subsec:AD-VAE}

As shown in Figure \ref{fig:model}, the latent variables $z$ is split into two parts: $z_y\in\mathbb{R}^L$ and $z_o\in\mathbb{R}^K$. We regard some face attributes as a part of variations of face images and specify a factored set of variables $z_y$ to capture these information. Each dimension of $z_y$ represents one single attribute. The remaining factors of variation, such as position and background, are captured by $z_o$. $z_y$ and $z_o$ are independent and the dimensions of $z_y$ are also independent with each other.

While the prior $p_\theta(z_o)$ remains to be an isotropic unit Gaussian, we choose a conditional distribution $p_\theta(z_y|y)$ as the prior of $z_y$ instead:
\begin{gather*}
	p_\theta(z_o)\sim{}N(0,I) \\
	p_\theta(z_y^i|y^i)\sim{}N(y^i,\sigma)	
\end{gather*}
where $y^i$ resfers to $i_{th}$ binary attribute label and $\sigma$ is the standard deviation of $p_\theta(z_y|y)$.  Then the variational lower bound described in Eq. \ref{eq:vae} can be rewritten as:
\begin{equation}
	\begin{split}
	L(x,y;\phi,\theta)={} &-\sum_{i=1}^L\alpha^i D_{KL}(q_\phi(z_y^i|x)||p_\theta(z_y^i|y^i)){} \\
						 &-\beta D_{KL}(q_\phi(z_o|x)||p_\theta(z_o)){} \\
						 &+E_{q_\phi(z_o|x)q_\phi(z_y|x)}[logp_\theta(x|z_o,z_y)]
	\end{split}
\end{equation}
where $q_\phi(z_o|x)$, $q_\phi(z_y|x)$ and $p_\theta(x|z_o,z_y)$ are multivariate Gaussian distributions parameterized by deep neural networks.


The first term of above formula is discriminative because we specify the mean of prior $p_{\theta}(z_y|y)$ to be the binary attribute label $y$ of the input image. Hence $q_{\phi}(z_y|x)$ also can serve as a classifier of face attributes by distinguishing the sign of the predicted mean of the posterior. In order to augment the quality of disentanglement with respect to the specified face attributes, we set a regularization coefficient vector $\alpha\in\mathbb{R}^L$ with large values to the discriminative $KL$ term of the attributes. Larger $\alpha$ will lead to higher classification accuracy but worse reconstruction fidelity. Following $\beta$-VAE \cite{Higgins2017β-VAE}, we also set another coefficient $\beta$ to the KL term of other latent variables to hinder these variables from encoding variations of face attributes. We choose a smaller value of  $\beta$ than $\alpha$.

\subsection{Photo Synthesis from Simple Line Drawing}

In order to sythesize face photo $x$ from simple line drawing $s$ controlled by face attributes $y$, we maximaze the variational lower bound of the conditinal log-likelihood $logp_{\theta}(x|s)$ instead:
\begin{equation}
	\begin{split}
	L(x,y,s;\phi,\theta)={} &-\sum_{i=1}^L\alpha^i D_{KL}(q_\phi(z_y^i|x)||p_\theta(z_y^i|y^i)){} \\
						    &-\beta D_{KL}(q_\phi(z_o|x)||p_\theta(z_o)){} \\
						    &+E_{q_\phi(z_o|x)q_\phi(z_y|x)}[logp_\theta(x|z_o,z_y,s)]
	\end{split}
\end{equation}

$z_o$ then mainly captures variations of backgrond and illumination and we regard these variations as the style of face photos. As shown in Figure \ref{fig:model}, another channel of convolutional neural network is added to the AD-VAE which takes simple line drawing as input and its feature maps are concatenated to the decoder.

\section{Experiments}
\label{sec:experiments}

\subsection{Dataset}
\label{ssec:subhead}

We conduct our experiments on CelebA dataset \cite{liu2015faceattributes}. CelebA consists of 202599 face images annotated with 40 binary attributes such as male, young, smiling, etc. FDoG \cite{kang2007coherent} filter is employed on CelebA to simulate the simple line drawing data. We binarize the synthetic simple line drawings with random thresholds on the training stage to avoid overfitting to the particular style. Both photos and simple line drawings are cropped and resized to $64\times64$. We use 182637 image pairs for training and the remaining 19962 pairs for testing. Among the training data, 10\% are used for cross-validation. 38 attributes are seleted without wearing necklace and wearing necktie.

We also test on ZJU-VIPA Line Drawing Face Database \cite{liang2012face} which is build on CUHK Face Sketch Database (CUFS) \cite{wang2009face}. 

\subsection{Attibute Manipulation}
\label{ssec:Attibute Manipulation}

\begin{figure*}[htbp]
\centering
   \includegraphics[width=380pt,keepaspectratio]{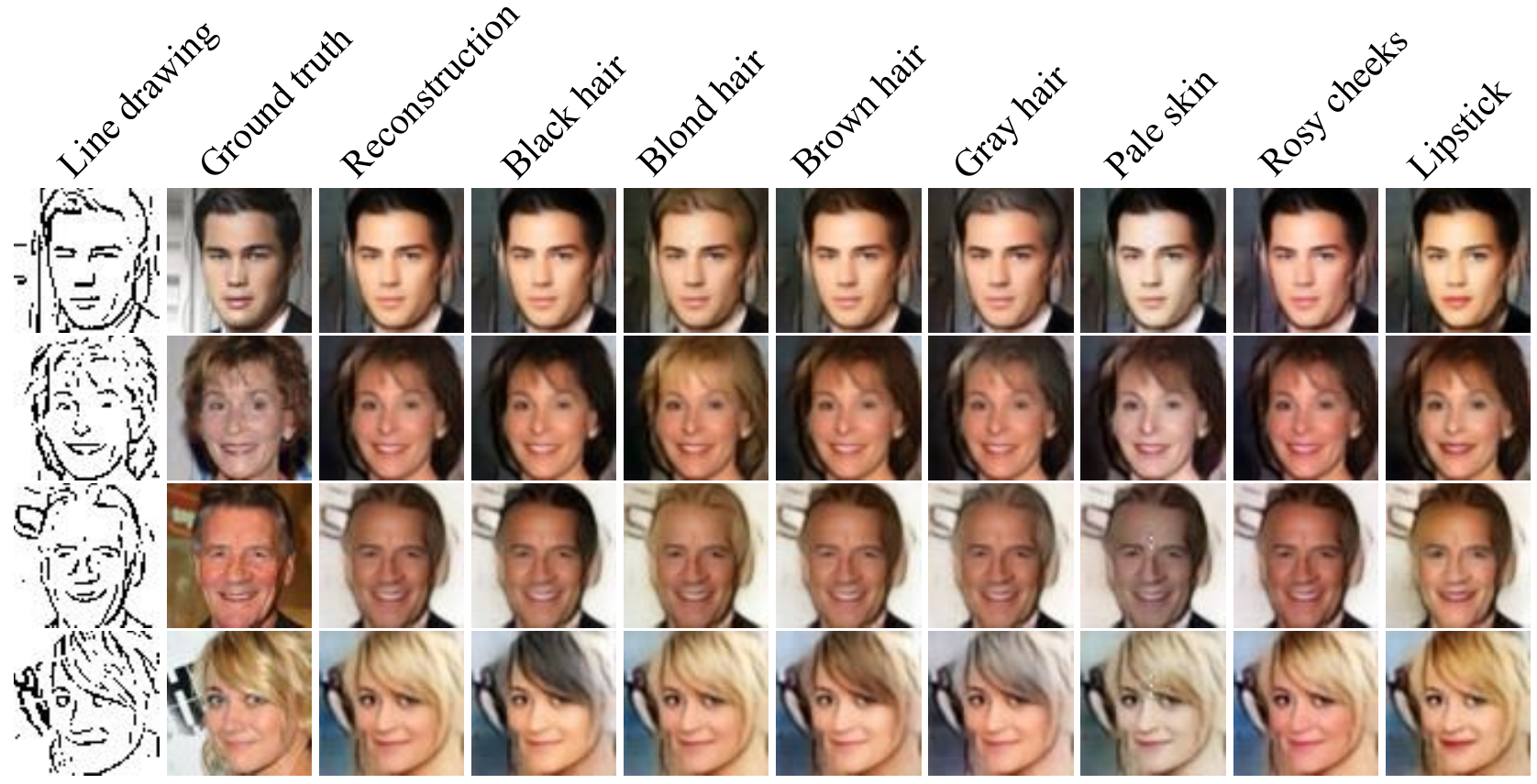}
   \caption{Attribute controlled photo synthesis}
   \label{fig:photo synthesis}
\end{figure*}

In order to demonstrate the qualitative disentanglement with respect to the face attributes, we manipulate the attribute variables by varying desired attribute variable smoothly and keeping all other latent variables fixed. We compare AD-VAE to attribute-conditioned variational auto-encoder (AC-VAE) \cite{yan2016attribute2image}. For fair model comparison, we train both AC-VAE and AD-VAE on CelebA with 38 selected binary attributes.


As shown in Figure \ref{fig:attribute}, with the desired attribute variable changes, the corresponding images generated by AD-VAE transform more visibly and naturally than those generated by AC-VAE. For example, when we increase the attribute variable of mouth\_slightly\_open, both two models generate faces with their mouths open, but AC-VAE uses teeth to fill the mouth no matter whether the teeth of ground truth are visible or not. While AC-VAE can hardly remove the sunglasses of input face, AD-VAE generates realistic eyes to replace the sunglasses. These results show our proposed AD-VAE can effectively separate the variations of specified attributes and other variations of face images.

\subsection{Photo Synthesis from Simple Line Drawing}
\label{ssec:Photo Synthesis from Simple Line Drawing}

\begin{figure}[htbp]
\centering
   \includegraphics[width=240pt, keepaspectratio]{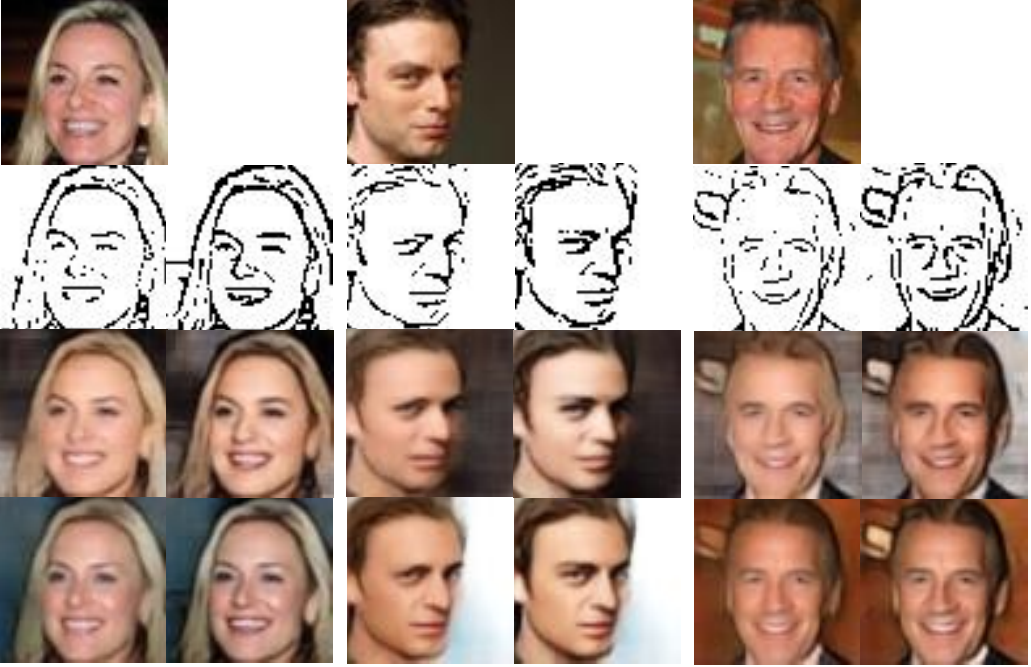}
   \caption{Results comparison. From top to bottom: ground truth, two types of simple line drawings, CSI, our method}
   \label{fig:different types}
\end{figure}

In this experiment, we choose some attributes not contained in simple line drawing to guide the photo synthesis. As shown in Figure \ref{fig:photo synthesis}, controlled by face attributes, we can modify the hair color and complexion of the synthesized face photos. In Figure \ref{fig:different types}, two types of simple line drawings with different stroke weights are used to generate face photos. Compared to the CSI \cite{guccluturk2016convolutional} with content loss, our proposed method can synthesize more photorealistic and natural faces images even though the contours of human faces are not complete. 

\begin{figure}[htbp]
\centering
   \includegraphics[width=220pt, keepaspectratio]{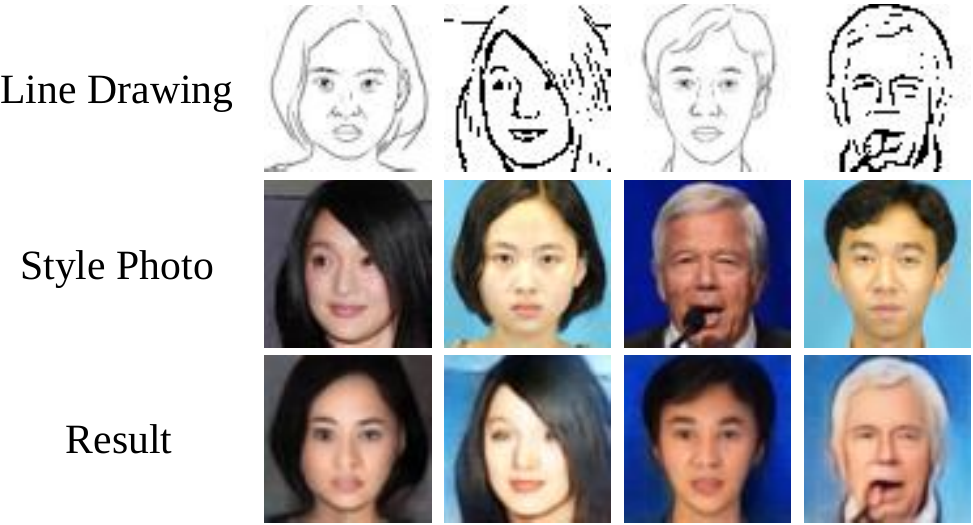}
   \caption{Target style photo synthesis}
   \label{fig:Target style photo synthesis}
\end{figure}

We exchange $z_o$ of several photos in ZJU-VIPA Line Drawing Face Database and CelebA dataset to synthesize face images with target styles in Figure \ref{fig:Target style photo synthesis}. As background and illumination are irrelevant to the human faces, this target style photo synthesis can eliminate these disturbances for further sketch-based face recogintion.

\section{Conclusion}
\label{sec:Conclusion}
This paper proposed a deep generative model to synthesize face photo from simple line drwing controlled face attributes. First, an attribute-disentangled variational auto-encoder (AD-VAE) is introduced to disentangle variations of face attributes from other variations of face images. Then we synthesized face photo from simple line drawing based on AD-VAE. Experiments showed our proposed method could learn interpretable representations of face images and generate face images with rich details and desired attributes even though the simple line drawing is not complete. We also did a target style photo synthesis that could help face recongnition and face verification in further step.

\bibliographystyle{IEEEbib}
\bibliography{citation}

\end{document}